\documentclass[10pt,twocolumn,letterpaper]{article}

\usepackage{iccv}
\usepackage{times}
\usepackage{epsfig}
\usepackage{graphicx}
\usepackage{amsmath}
\usepackage{amssymb}
\usepackage{multirow}
\usepackage{bm}
\usepackage{makecell}
\usepackage[ruled,linesnumbered]{algorithm2e}
\usepackage[flushleft]{threeparttable}
\usepackage{lipsum}

\usepackage[breaklinks=true,bookmarks=false]{hyperref}
\usepackage[accsupp]{axessibility}  % Improves PDF readability for those with disabilities
\newcommand\blfootnote[1]{%
  \begingroup
  \renewcommand\thefootnote{}\footnote{#1}%
  \addtocounter{footnote}{-1}%
  \endgroup
}

\iccvfinalcopy

\def\iccvPaperID{6926}

\ificcvfinal\fi

\begin{document}

%%%%%%%%% TITLE
\title{LoLep: Single-View View Synthesis with Locally-Learned Planes and Self-Attention Occlusion Inference}

\author{Cong Wang$^{1}$, Yu-Ping Wang$^{2}$, Dinesh Manocha$^{3}$ \\ \\
$^{1}$Tsinghua University, $^{2}$The Beijing Institute of Technology, $^{3}$University of Maryland
}

\maketitle
\ificcvfinal\thispagestyle{empty}\fi

\begin{abstract}
We propose a novel method, \textbf{\emph{LoLep}}, which regresses \textbf{\emph{Lo}}cally-\textbf{\emph{Le}}arned \textbf{\emph{p}}lanes from \textbf{a single RGB image} to represent scenes accurately, thus generating better novel views.
Without the depth information, regressing appropriate plane locations is a challenging problem.
To solve this issue, we pre-partition the disparity space into bins and design a disparity sampler to regress local offsets for multiple planes in each bin. 
However, only using such a sampler makes the network not convergent; we further propose two optimizing strategies that combine with different disparity distributions of datasets and propose an occlusion-aware reprojection loss as a simple yet effective geometric supervision technique.
We also introduce a self-attention mechanism to improve occlusion inference and present a Block-Sampling Self-Attention (BS-SA) module to address the problem of applying self-attention to large feature maps.
We demonstrate the effectiveness of our approach and generate state-of-the-art results on different datasets. 
Compared to MINE, our approach has an LPIPS reduction of 4.8\%$\sim$9.0\% and an RV reduction of 74.9\%$\sim$83.5\%.
We also evaluate the performance on real-world images and demonstrate the benefits. 
% We will release the source code at the time of publication.
\end{abstract}

%%%%%%%%% BODY TEXT
\section{Introduction}
\label{sec: introduction}

Single-view view synthesis allows a camera to roam around a scene from a given photograph. It has been used to generate compelling views for different applications including image editing and augmented or virtual reality.
The underlying techniques require understanding the geometry of scenes, reasoning about occlusions, and rendering high-quality images of novel views in real time.\blfootnote{Yu-Ping Wang is the corresponding author (wyp\_cs@bit.edu.cn).}

Many approaches have been proposed to solve this problem~\cite{DBLP:conf/eccv/XieGF16, DBLP:conf/iccv/SrinivasanWSRN17, DBLP:conf/cvpr/LiuHS18, DBLP:journals/tog/NiklausMYL19, DBLP:conf/cvpr/WilesGS020}.
They synthesize novel views by predicting a naive representation (e.g., depth maps, voxels, or point clouds) from a single image and generating images for novel views using appropriate rendering techniques.
While these methods generate some positive results, they limit the performance of single-view view synthesis due to their inability to represent occluded regions well~\cite{DBLP:conf/cvpr/TuckerS20}.
In this context, layered representations~\cite{DBLP:conf/cvpr/TuckerS20, DBLP:conf/eccv/TulsianiTS18, DBLP:journals/tog/ZhouTFFS18, DBLP:journals/tog/LiK20, DBLP:conf/iccv/LiFSDWL21, DBLP:conf/siggraph/HanWY22} are more suitable for single-view view synthesis.

\begin{figure}
    \centering
    \includegraphics[scale=0.61]{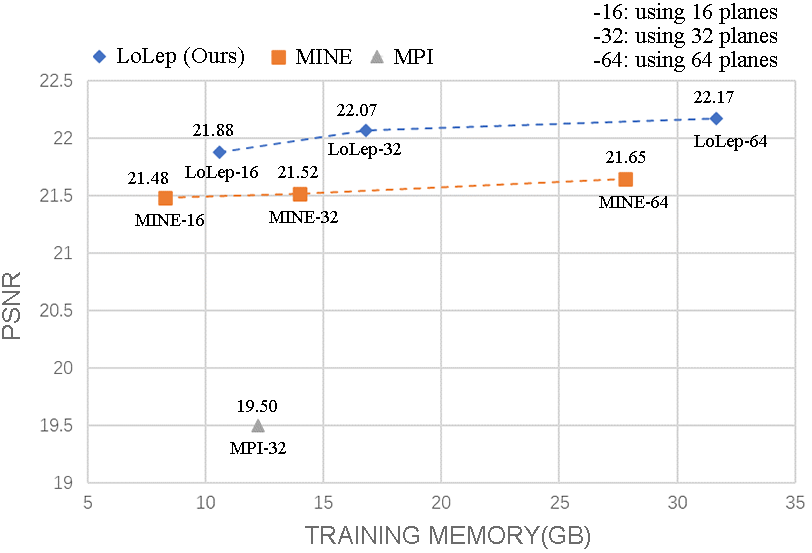}
    \caption{{\bf Comparisons on the KITTI dataset.}
    LoLep generates state-of-the-art results and even LoLep with fewer planes uses less memory and generates better novel views than previous methods with more planes (LoLep-16 vs. MINE-32, MINE-64 and MPI-32, LoLep-32 vs. MINE-64), which benefits from locally-learned planes and self-attention occlusion inference.
    The batch size is 4.
    }
    \label{fig: teaser}
\end{figure}

\begin{figure*}
    \includegraphics[scale=0.088]{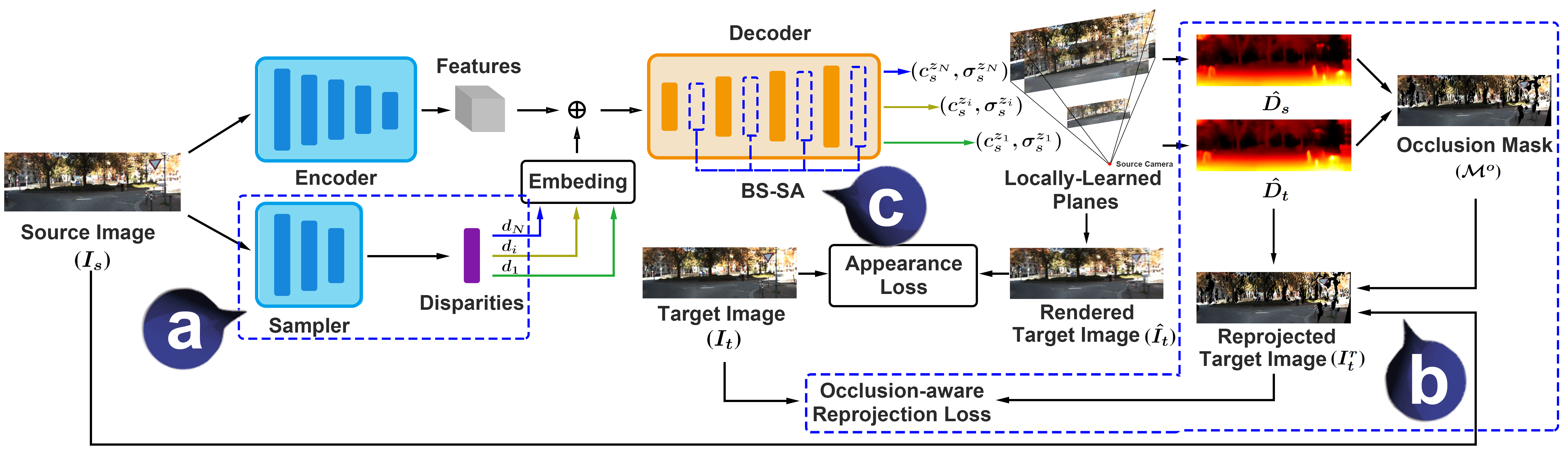}
    \caption{{\bf Overview.}
    LoLep regresses locally-learned planes to represent scenes accurately without a depth map input mainly relying on three novel components.
    {\bf (a) Disparity Sampler:} regressing accurate locations for multiple planes from only the RGB image; 
    {\bf (b) Occlusion-aware Reprojection Loss:} a simple yet effective geometric supervision technique for single-view view synthesis to learn better geometry;
    {\bf (c) Block-Sampling Self-Attention:} supporting self-attention applied to large feature maps for higher performance.
    `$\oplus$' concatenates two tensors.
    }
    \label{fig: pipeline}
\end{figure*}

Recently, Multiplane Image (MPI)~\cite{DBLP:journals/tog/ZhouTFFS18} has gained popularity as a layered representation and has been used for single-view view synthesis~\cite{DBLP:conf/cvpr/TuckerS20}.
Specifically, it is an encoder-decoder structure supervised by multiple images from different views of a given scene and is used to predict multiple planes of RGB and alpha values from a single image.
MINE~\cite{DBLP:conf/iccv/LiFSDWL21} combines MPI with NeRF~\cite{DBLP:conf/eccv/MildenhallSTBRN20} and generalizes MPI into a continuous depth MPI by considering multiple plane location inputs. 
This can improve the performance of single-view view synthesis to better infer geometric primitives in a scene.
However, these methods sample plane locations randomly, which makes it hard for the planes to learn optimal scene representations.
As a result, these methods usually require more planes to obtain satisfactory novel views, requiring huge computing power.
To alleviate this requirement, a key issue is \emph{how to fully utilize the limited planes to obtain the most accurate scene representation as possible}.

\begin{table*}[t]
    \caption{Symbols and Notation}
    \centering
    \resizebox{160mm}{!}{
    \begin{tabular}{|c|l|c|l|c|l|}
    \hline
    $I$ & image & $d_f$ & far disparity & $(\cdot)_s$ & an entity related to the source view or the source camera \\ \hline
    $D$ & depth map & $d_n$ & near disparity & $(\cdot)_t$ & an entity related to the target view or the target camera \\ \hline
    $d$ & disparity value & $N$ & number of planes & $(\cdot)_i$ & an entity related to the i-th plane \\ \hline
    $z$ & depth value & $K$ & intrinsic matrix & $(\cdot)^{z_i}$ & an entity related to the plane, whose depth is $z_i$  \\ \hline
    $H$ & height & $T$ & transformation matrix & $\theta_{ED}$ & parameters of the encoder-decoder \\ \hline
    $W$ & width & $R$ & rotation matrix & $\theta_{S}$ & parameters of the disparity sampler \\ \hline
    $c$ & RGB values & $t$ & translation vector & $\pi(\cdot)$ & projecting a 3D camera coordinate to 2D pixel coordinate \\ \hline
    $\sigma$ & volume density values & $\hat{(\cdot)}$ & predicted results & $\pi^{-1}(\cdot)$ & the inverse operation of $\pi(\cdot)$ \\ \hline
    \end{tabular}
    }
    \label{tab: symbols}
\end{table*}

Previous works~\cite{DBLP:journals/tog/LiK20, DBLP:conf/siggraph/HanWY22} solve this issue by regressing more accurate locations for multiple planes. 
Due to the lack of supervision and using globally-learned planes, however, their networks take an RGB image and an additional depth map as input.
The depth map is provided by a pretrained depth prediction network, which introduces a heavy dependence on other networks.

\noindent {\bf Main Results:} 
We present a novel single-view view synthesis method based on Multiplane Image, LoLep.
LoLep aims to make full use of locally-learned planes to represent scenes accurately, thus generating better novel views from a single RGB image with less memory (Figure ~\ref{fig: teaser}).
In order to achieve that, we pre-partition the disparity space into bins and design a disparity sampler to condition local offsets of planes on a single RGB image.
However, due to the lack of depth information, applying the sampler directly makes the network not convergent.
We further propose two optimizing strategies that combine with different disparity distributions of datasets and an occlusion-aware reprojection loss to solve it (described in Section \ref{sec: LLP}).
To improve the ability for occlusion inference, we introduce a self-attention mechanism to our decoder and present a Block-Sampling Self-Attention (BS-SA) module to work for large feature maps (described in Section \ref{sec: MSA}).
Overall, the novel components of our approach include:

\begin{itemize}
    \setlength{\itemsep}{0pt}
    \setlength{\parsep}{0pt}
    \setlength{\parskip}{1pt}
    \item We propose a novel single-view view synthesis method based on Multiplane Image, LoLep, that regresses accurate scene representations and generates better novel views on scene geometry and occluded regions.
    \item We introduce a self-attention mechanism to improve occlusion inference and present a BS-SA module to address the  problem  of  applying  self-attention on large  feature  maps.
    \item We compare with prior methods and show that LoLep outperforms MINE on different datasets with an LPIPS reduction of 4.8\%$\sim$9.0\% and an RV reduction of 74.9\%$\sim$83.5\%. 
    Moreover, LoLep with fewer planes uses less memory and generates better results than prior methods with more planes.
\end{itemize}

\section{Related Works}

{\bf Multi-view View Synthesis.}
Multi-view view synthesis is a well-studied problem, and methods in this area generate images for novel views given a set of images from different views of the same scene. 
Some earlier methods are based on interpolating nearby views~\cite{DBLP:conf/siggraph/LevoyH96, DBLP:conf/siggraph/GortlerGSC96, DBLP:conf/siggraph/BuehlerBMGC01}. 
However, synthesizing novel views using interpolation techniques without a 3D representation causes inconsistency between different generated views.
To alleviate this problem, many approaches based on depth maps~\cite{DBLP:journals/tog/ChaurasiaDSD13, DBLP:journals/tog/PennerZ17} and multi-view geometry~\cite{DBLP:conf/siggraph/DebevecTM96, DBLP:conf/iccv/FitzgibbonWZ03, DBLP:journals/tog/ZitnickKUWS04, DBLP:journals/tog/KopfLSSG13, DBLP:conf/icra/WangWM22, DBLP:conf/iros/WangZWDQM21} have been proposed.
Moreover, deep learning methods have also been applied to novel view synthesis~\cite{ DBLP:journals/tog/ZhouTFFS18, DBLP:journals/tog/HedmanPPFDB18, DBLP:conf/cvpr/SitzmannTHNWZ19, DBLP:conf/iccv/ChoiGT0K19, DBLP:conf/cvpr/MeshryGKHPSM19, DBLP:conf/eccv/AlievSKUL20, DBLP:conf/siggrapha/WangKCBSZ23, DBLP:conf/cvpr/WangKSQWBZ25, DBLP:conf/siggrapha/WangLBXZW19}.
Some of these methods use deep neural networks to improve on traditional approaches, so they can be applied to more challenging scenarios.
Recently, NeRF~\cite{DBLP:conf/eccv/MildenhallSTBRN20} techniques have been used to generate improved results for view synthesis.
However, these techniques can only generate novel views of specific static scenes and involve intensive computation.
Many techniques have also been proposed to improve the performance~\cite{DBLP:journals/corr/abs-2010-04595, DBLP:conf/cvpr/YuYTK21, DBLP:conf/cvpr/Martin-BruallaR21,DBLP:conf/iccv/ParkSBBGSM21, DBLP:journals/corr/abs-2106-05264, DBLP:journals/corr/abs-2204-00928, DBLP:journals/corr/abs-2508-09597}.
Unlike these methods, we deal with the problem of single-view view synthesis, which only has one input image at test time.

{\bf Single-view View Synthesis.}
Compared to multi-view view synthesis, single-view view synthesis is a more challenging task and has wider applications.
Deep3D~\cite{DBLP:conf/eccv/XieGF16} predicts a probabilistic disparity map from the left eye's view and renders a novel image for the right eye. 
The missing regions are inpainted implicitly using neural networks.
\cite{DBLP:journals/tog/NiklausMYL19} renders novel views using a predicted depth map. 
They utilize context-aware inpainting to fill in missing regions, thereby generating better results.
To make single-view view synthesis more general, SynSin~\cite{DBLP:conf/cvpr/WilesGS020} takes a single image as input and can synthesize an image at any given pose.
Recently, many approaches based on layered representations have been proposed that are better at handling occluded regions. 
\cite{DBLP:conf/eccv/TulsianiTS18} uses layered depth images (LDI) for single-view view synthesis, successfully inferring not only the depth of visible pixels but also the texture and depth of content that is occluded.
%Given the success of Zhou et al.~
\cite{DBLP:conf/cvpr/TuckerS20} performs single-view view synthesis using  MPI, which results in better performance.
% than~\citet{DBLP:conf/eccv/TulsianiTS18}.
\cite{DBLP:journals/tog/LiK20} extends MPI representation and proposes Variable MPI, which allows the locations of multiple planes to be inferred from the input image and the depth map.
To further explore the potential of MPI for view synthesis, \cite{DBLP:conf/iccv/LiFSDWL21} combines MPI with NeRF and propose a novel layered representation called MINE, which results in considerable performance improvement.
AdaMPI~\cite{DBLP:conf/siggraph/HanWY22} was recently proposed to synthesize novel views for in-the-wild photographs, but it still requires a depth map input.
In general, these prior MPI-based methods either randomly sample plane locations, which requires more planes and incurs huge computing overhead, or learn plane locations globally, which requires an additional depth map input.

\section{Background}
\label{sec: preliminary}

The symbols and notation in this paper are defined in Table \ref{tab: symbols}.
Our approach employs the same scene representation as MINE~\cite{DBLP:conf/iccv/LiFSDWL21}, which is referred to as \emph{MINE planes} in this paper.
MINE planes are a set of 4-channel (i.e., RGB and volume density) planes parallel to the current camera at different disparities and can be used to render the image and depth map in the current view using volume rendering.
Specifically, given MINE planes (i.e., $\{c^{z_i}, \sigma^{z_i}|i=1 \cdots N\}$), the current view can be rendered as:
\begin{equation}
\label{eq: render}
    \textstyle{\hat{I} = \sum_{i=1}^N w_i c^{z_i}, w_i = \mathcal{T}_i (1 - \exp(- \sigma^{z_i} \delta^{z_i}))},
    % \hat{I} = \sum_{i=1}^N w_i c^{z_i}, w_i = \mathcal{T}_i (1 - \exp(- \sigma^{z_i} \delta^{z_i})),
\end{equation}
where $\mathcal{T}_i = \exp(- \sum_{j=1}^{i-1} \sigma^{z_j} \delta^{z_j}) : \mathbb{R}^2 \rightarrow \mathbb{R}^+$ is the map of accumulated transmittance from the first plane to the $i$-th plane, and $\mathcal{T}_i(x, y)$ denotes the probability of a ray traveling from $(x, y, z_1)$ to $(x, y, z_i)$ without hitting any object. Furthermore, $\delta^{z_i}(x, y) = || \pi^{-1}([x, y, z_{i+1}]^T) - \pi^{-1}([x, y, z_i]^T) ||_2 : \mathbb{R}^2 \rightarrow \mathbb{R}^+ $ is the distance map between planes $i+1$ and $i$. The depth map of the current view can be rendered similar to Eq. (\ref{eq: render}), i.e.:
\begin{equation}
\label{eq: d_render}
    \textstyle{\hat{D} = \sum_{i=1}^N w_i z_i}.
    % \hat{D} = \sum_{i=1}^N w_i z_i.
\end{equation}

Given MINE planes in the source view, a target view can be generated using a homography warping and volume rendering~\cite{DBLP:conf/siggraph/KajiyaH84, DBLP:conf/eccv/MildenhallSTBRN20}. 
Following the standard inverse homography~\cite{DBLP:conf/cvpr/TuckerS20, DBLP:books/daglib/0015576, DBLP:journals/tog/ZhouTFFS18}, the correspondence between a pixel coordinate $[x_s, y_s]^T$ in a source plane and a pixel coordinate $[x_t, y_t]^T$ in a target plane is given by: 
\begin{equation}
\label{eq: corresponse}
    \textstyle{[x_s, y_s, 1]^T \sim K_s ( R - \frac{tn^T}{z_i} ) K_t^{-1} [x_t, y_t, 1]^T},
    % [x_s, y_s, 1]^T \sim K_s ( R - \frac{tn^T}{z_i} ) K_t^{-1} [x_t, y_t, 1]^T,
\end{equation}
where $n = [0, 0, 1]^T$ is the normal vector of MINE planes in the target view.
For brevity, we denote Eq. (\ref{eq: corresponse}) as $[x_s, y_s]^T = \mathcal{W}(x_t, y_t)$.
MINE planes in the source view can then be projected to the target view as: $c_t^{z_i}(x_t, y_t) = c_s^{z_i}(\mathcal{W}(x_t, y_t))$, $\sigma_t^{z_i}(x_t, y_t) = \sigma_s^{z_i}(\mathcal{W}(x_t, y_t))$.
Generated MINE planes in the target view are used to render the target image and the target depth map using Eqs. (\ref{eq: render}) and (\ref{eq: d_render}).

\section{Our Method}
In this section, we present our novel approach for single-view view synthesis.
Figure \ref{fig: pipeline} gives an overview of our approach.
The source image is first fed into the encoder and the disparity sampler.
The encoder is used to extract image features, and the disparity sampler is used to regress locally-learned plane locations (described in Section \ref{sec: sampler}).
The regressed locations are embedded in the same manner as NeRF~\cite{DBLP:conf/eccv/MildenhallSTBRN20} and concatenated with extracted features through channels.
The decoder takes the concatenated features as input and predicts locally-learned planes in the source view, which can be used to render images in novel views.
The appearance loss is computed mainly using the difference between the ground truth and predicted novel views.
To build our occlusion-aware reprojection loss, an occlusion mask is first obtained using our detection method, and the reprojection loss is computed as the masked difference between the projected image and the ground truth (described in Section \ref{sec: reprojection}).
Our BS-SA module can be applied after any layer of the decoder to handle occlusions without worrying about large feature maps (described in Section \ref{sec: MSA}).

\subsection{Locally-Learned Planes} 
\label{sec: LLP}

\subsubsection{The Point for Locally-Learned Planes}
\label{sec: point}

\textbf{Compared to fixed planes.}
The insight on using MINE planes is to approximate the integral of volume rendering using the rectangular approximation method. 
Imagine that there are two rays that intersect on a pixel of a given plane.
If the plane location is fixed, the network can only set this pixel to the average of densities of sampled points in two views, aiming to obtain a good approximation of the integral in both views.
However, if the plane location is flexible (i.e., learned), the network can find a better plane location at which the densities of sampled points in two views are more similar or exactly the same, which provides a more accurate approximation of the integral of volume rendering for all views.

\noindent
\textbf{Compared to globally-learned planes.}
Since networks tend to produce low-frequency outputs, if there is not enough supervision or regularization, globally-learned planes would cluster around a certain disparity, and one of those planes would cluster all rendering weights (shown in our supplementary materials).
Therefore, previous methods with globally-learned planes usually require a depth map as an additional input~\cite{DBLP:journals/tog/LiK20, DBLP:conf/siggraph/HanWY22}.
Locally-learned planes itself as a regularization can avoid this issue.

For the reasons above, we first propose a disparity sampler to regress plane locations, which is then fed into a decoder to obtain locally-learned planes.
However, a direct application of such a pipeline still makes the network not convergent due to the lack of depth information.
We further propose two optimizing strategies that combine different disparity distributions of datasets to solve this issue.
In addition, an occlusion-aware reprojection loss is also explored as a novel geometric supervision technique for single-view view synthesis.

\subsubsection{Disparity Sampler}
\label{sec: sampler}
We design the disparity sampler as an encoder, taking a single image as input and regressing several offsets $\{v_i|0 < v_i < 1, i=1 \cdots N\}$. 
For locally learning, we pre-partition the disparity space $[d_f, d_n]$ into N bins uniformly, and the locations of locally-learned planes are computed as:
\begin{equation}
\label{eq: sampler}
    % \textstyle{d_i = s_i + v_i l_i}.
    \textstyle{d_i = d_n + (v_i + i - 1) \frac{d_f - d_n}{N}}.
    % d_i = d_n + (v_i + i - 1) \frac{d_f - d_n}{N}.
\end{equation}
Our formulation naturally restricts each locally-learned plane into the corresponding bin, thereby preventing planes from clustering as globally-learned planes do.

We observe that different datasets may have different disparity distributions, which impacts the convergence of our network.
We divide these disparity distributions into two cases.
The first is \emph{the uniform disparity distribution}, which has approximately the same number of pixels at each disparity, while the second is \emph{the aggregated disparity distribution}, which has most pixels concentrated at some disparities that are far apart and only a few pixels at the rest of the values.
Detailed descriptions and visualizations of the distributions can be found in our supplementary materials. 
To make our disparity sampler work well with both disparity distributions, we propose the following parameter optimizing strategies.

{\bf Parameter optimizing strategy for uniform disparity distribution (U-opt):} 
For images with uniform disparity distributions (e.g., the KITTI and RealEstate10K datasets), there are enough pixels in each bin to optimize the network parameters. Therefore, we propose U-opt to simultaneously optimize $\theta_{ED}$ and $\theta_S$ to fit $(\bm{c}, \bm{\sigma}) = \mathcal{F}_{\theta_{ED}, \theta_{S}}(I_s, \mathcal{S}(I_s))$.

\begin{figure*}
    \centering
    \includegraphics[scale=0.21]{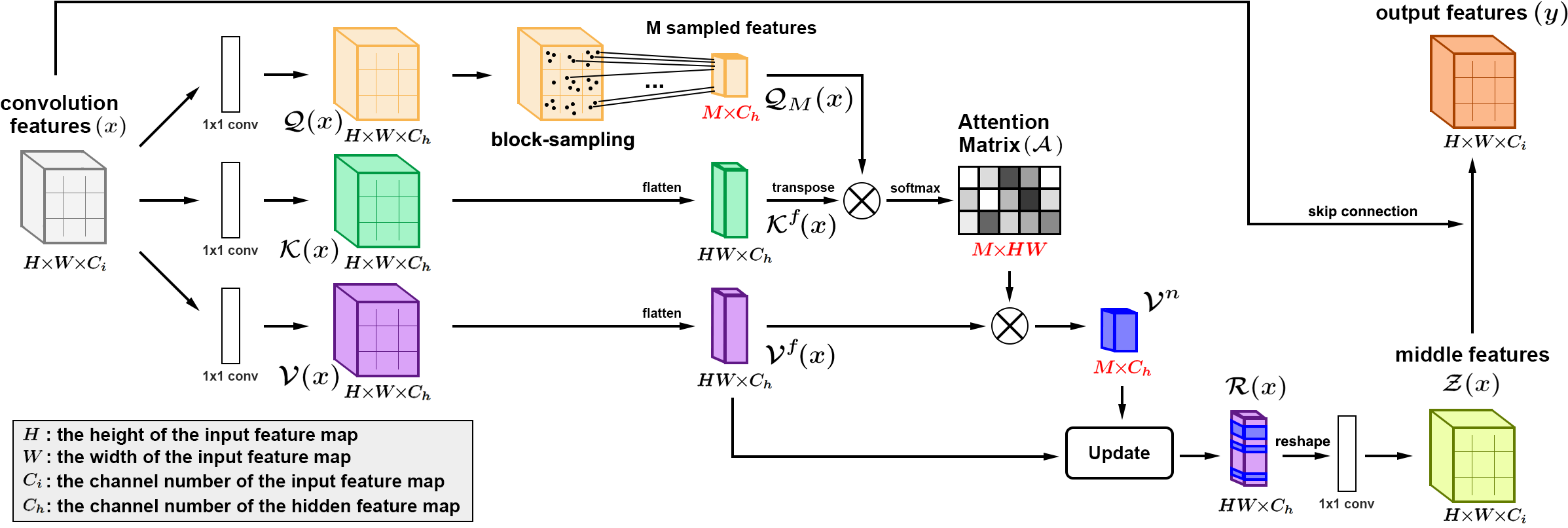}
    \caption{{\bf Block-Sampling Self-Attention Module.}
    The block-sampling self-attention module reduces the size of the attention matrix from $HW \times HW$ to $M \times HW$ and solves the issue that the original self-attention mechanism cannot be applied to large feature maps.
    $M$ is a hyper-parameter.
    ``$\otimes$'' denotes matrix multiplication.
    The softmax operation is performed on each row.
    }
    \label{fig: MSA}
\end{figure*}

{\bf Parameter optimizing strategy for aggregated disparity distribution (A-opt):} 
For images with aggregated disparity distributions (e.g., the Flowers Light Field dataset), there could be only a few pixels in some bins for optimization, which cannot provide enough supervision to learn $\theta_{ED}$ and $\theta_S$ jointly.
Therefore, we design A-opt, which uses a two-stage procedure.
In the first stage, we optimize $\theta_{ED}$ without the disparity sampler.
In the second stage, we employ the full pipeline in Figure \ref{fig: pipeline}, learning $\theta_{ED}$ with a small learning rate and  $\theta_S$ with a big one.
The first stage aims to provide a better initialization for the encode-decoder, based on which the sampler can be updated in the right direction during the second stage even with a few pixels.

Our proposed sampler is somewhat similar to Adabins~\cite{DBLP:conf/cvpr/BhatAW21}.
Adabins and our method both attempt to get a depth distribution prior for each image, thus obtaining better prediction.
However, due to different tasks and conditions, we have different designs:
(1) With the ground truth depth, Adabins learns depth distributions globally, which is not feasible for our task, as explained in Section~\ref{sec: point}.
(2) The task of Adabins is the monocular depth estimation, so Adabins employ 
a heavy network (an encoder-decoder and an mViT) to obtain a depth distribution prior (compared to our disparity sampler).
This will cause high computing requirements, which is not expected in our work.

\subsubsection{Occlusion-Aware Reprojection Loss}
\label{sec: reprojection}
The occlusion-aware reprojection loss supervises only rendered depth maps, making up for the lack of depth supervision and helping to obtain better scene geometry.
According to multi-view geometry~\cite{DBLP:books/daglib/0015576}, a pixel coordinate in the target image $[x_t, y_t]^T$ can be projected to a camera coordinate in the source view $[X_s, Y_s, Z_s]^T$ as
\begin{equation}
\label{ref: tgt_bak_project}
    \textstyle{[X_s, Y_s, Z_s]^T = T_{t \rightarrow s} \pi^{-1}([x_t, y_t, \hat{D_t}(x_t, y_t)]^T)}.
    % [X_s, Y_s, Z_s]^T = T_{t \rightarrow s} \pi^{-1}([x_t, y_t, \hat{D_t}(x_t, y_t)]^T).
\end{equation}
$[X_s, Y_s, Z_s]^T$ can be further projected to a pixel coordinate $[x_s, y_s]^T$ in the source image as
\begin{equation}
\label{ref: src_project}
    \textstyle{[x_s, y_s]^T = \pi ([X_s, Y_s, Z_s]^T)}.
    % [x_s, y_s]^T = \pi ([X_s, Y_s, Z_s]^T),
\end{equation}
Then a pixel $[x_t, y_t]^T$ in the target image is considered occluded if $Z_s - \hat{D_s}(x_s, y_s) >= c \cdot s$, 
where $c$ is a constant that equals 0.2 in our experiments and $s$ is the scale of learned planes.
The generated occlusion mask is denoted as $\mathcal{M}^o$, with 1 for occluded pixels and 0 for others.

Based on $\mathcal{M}^o$, the occlusion-aware reprojection loss can be computed as:
\begin{equation}
    \textstyle{L_{rep} = \frac{1}{HW} \sum |I_t - I_t^r| \cdot (\bm{1} - \mathcal{M}^o)},
    % L_{rep} = \frac{1}{HW} \sum |I_t - I_t^r| \cdot (\bm{1} - \mathcal{M}^o),
\end{equation}
where $I_t$ is the ground truth in the target view. 
$I_t^r$ is the image in the target view projected from the ground truth in the source view using Eqs. (\ref{ref: tgt_bak_project}) and (\ref{ref: src_project}).

Combined with the reprojection loss, our overall loss is:
\begin{equation}
    \textstyle{L_{total} = L_{app} + \lambda L_{rep}},
\end{equation}
where $L_{app}$ is the appearance loss, built on the edge-aware smoothness loss~\cite{DBLP:conf/cvpr/GodardAB17, DBLP:conf/iccv/GodardAFB19} and the L1 loss between $\hat{I_t}$ and $I_t$. $\lambda$ is set to 1 after searching in a manual range.

\subsection{Self-Attention Occlusion Inference}
\label{sec: MSA}

The self-attention mechanism~\cite{DBLP:conf/cvpr/0004GGH18} improves the performance of neural networks by considering the correlation between features.
Intuitively, it can be employed to infer occluded pixels using dis-occluded regions.
However, due to the huge size of the attention matrix, the self-attention mechanism has prohibitive computational cost and vast video memory occupation~\cite{DBLP:conf/iccv/ZhuXBHB19} and is hard to be used on large feature maps for higher performance.
To alleviate this problem, we propose a BS-SA module.

As shown in Figure \ref{fig: MSA}, image features from the previous layer $\bm{x} \in \mathbb{R}^{H \times W \times C_i}$ are first transformed into different feature spaces $\mathcal{Q}(x) \in \mathbb{R}^{H \times W \times C_h}$, $\mathcal{K}(x) \in \mathbb{R}^{H \times W \times C_h}$, and $\mathcal{V}(x) \in \mathbb{R}^{H \times W \times C_h}$ using $1 \times 1$ convolutions.
Unlike the original self-attention mechanism that causes an attention matrix of size $HW \times HW$, our BS-SA module reduces the size to $M \times HW$ with slight accuracy sacrifice by block-sampling M query points during each training step.
Specifically, during each training step, we block-sample M locations in feature maps and take features of $\mathcal{Q}(x)$ at these locations as the query vector instead of all the features of $\mathcal{Q}(x)$.
The query vector is then multiplied with flattened features of $\mathcal{K}(x)$ to obtain a smaller attention matrix $\mathcal{A}$. 
The resulting features of query points can be computed by multiplying $\mathcal{A}$ with the flattened features of $\mathcal{V}(x)$, while those of other points are set to the same as $\mathcal{V}(x)$. 
We summarize our BS-SA module in Algorithm \ref{alg: BS-SA}.

\begin{algorithm}[]
	\caption{Block-Sampling Self-Attention.}
	\label{alg: BS-SA}
	\KwIn{
	The features from the previous layer $\bm{x}$;
	The number of sample points $M$.
	}
	\KwOut{Output features $\bm{y}$.}  
	\BlankLine

    Taking $\bm{x}$ as input, compute $\mathcal{Q}(\bm{x}), \mathcal{K}(\bm{x}),$ and $\mathcal{V}(\bm{x})$ using 1x1 convolutions.
    
    Randomly block-sample $M$ features in $\mathcal{Q}(\bm{x})$, and take sampled features as the query vector $\mathcal{Q}_M(\bm{x})$.
    
    $\mathcal{K}^f(\bm{x}) \leftarrow \textnormal{flatten}(\mathcal{K}(\bm{x}))$;
    
    $\mathcal{V}^f(\bm{x}) \leftarrow \textnormal{flatten}(\mathcal{V}(\bm{x}))$;
    
    $\mathcal{A} \leftarrow \textnormal{softmax}(\mathcal{Q}_M(\bm{x}) \times \mathcal{K}^f(\bm{x})^T$);
    
    $\mathcal{V}^{new} \leftarrow \mathcal{A} \times \mathcal{V}^f(\bm{x})$;
    
    Update $\mathcal{V}^f(\bm{x})$ using $\mathcal{V}^{new}$ to obtain the resulting features $\mathcal{R}(\bm{x})$;
    
    Taking $\mathcal{R}(\bm{x})$ as input, compute middle features $\mathcal{Z}(\bm{x})$ using a 1x1 convolution;
    
    $\bm{y} \leftarrow \mathcal{Z}(\bm{x}) + \bm{x}$.
\end{algorithm}

\section{Implementation and Results}
\label{sec: experiments}

In this section, we describe the implementation and evaluate the performance on different datasets.
We perform both quantitative and qualitative comparisons on the KITTI~\cite{DBLP:journals/ijrr/GeigerLSU13}, Flowers Light Fields~\cite{DBLP:conf/iccv/SrinivasanWSRN17}, and RealEstate10K~\cite{DBLP:journals/tog/ZhouTFFS18} datasets. We use the same metrics (SSIM, PSNR, and LPIPS
\footnote{
\label{foot: lpips}
The reported results on LPIPS in MINE~\cite{DBLP:conf/iccv/LiFSDWL21} is wrong. 
They input images in the range
[0, 1] to the LPIPS function, instead of [-1, 1].
(\url{https://github.com/vincentfung13/MINE/issues/4}).}
)
as previous works~\cite{DBLP:conf/cvpr/TuckerS20, DBLP:conf/iccv/LiFSDWL21} to measure the quality of synthesized images and propose a new metric, Rendering Variance (RV), to measure the dispersion of weights in the volume rendering.
We conduct many  ablation studies on the KITTI dataset to demonstrate the functionality of each proposed component, and a depth evaluation is also performed on the NYU-Depth V2~\cite{DBLP:conf/eccv/SilbermanHKF12} and iBims-1~\cite{DBLP:conf/eccv/KochLF018} datasets.
Moreover, we compare our model with MINE on real-world images in our supplementary materials (SMs).

\subsection{Rendering Variance}
\label{sec: RV}

Given a ray $\phi$ with N sample points, rendering variance (RV) is formulated as:
\begin{equation}
    \textstyle{RV(\phi) = \sum_i w_i (s \cdot z_i - z)^2.}
\end{equation}
$w_i$ is defined in Eq. (\ref{eq: render}), and $z_i$ is the depth of the $i$-th sample.
$z$ is the ground truth depth.
$s$ is a relative scale to solve the scale ambiguity of depth from a single image~\cite{DBLP:conf/cvpr/TuckerS20}.
RV computes a weighted variance of depths of the sample points, and the weights are obtained using volume rendering.
Intuitively, with smaller RVs, the rendering will concentrate on fewer sample points around the real depth, which may not do much for the current view but will generate sharper images with fewer artifacts for novel views.
Since the RealEstate10K and Flowers Light Field dataset do not provide the ground truth depth, we only compare RV on the KITTI dataset in our experiments (using their public LiDAR data).

\begin{figure*}
\centering
\includegraphics[scale=0.378]{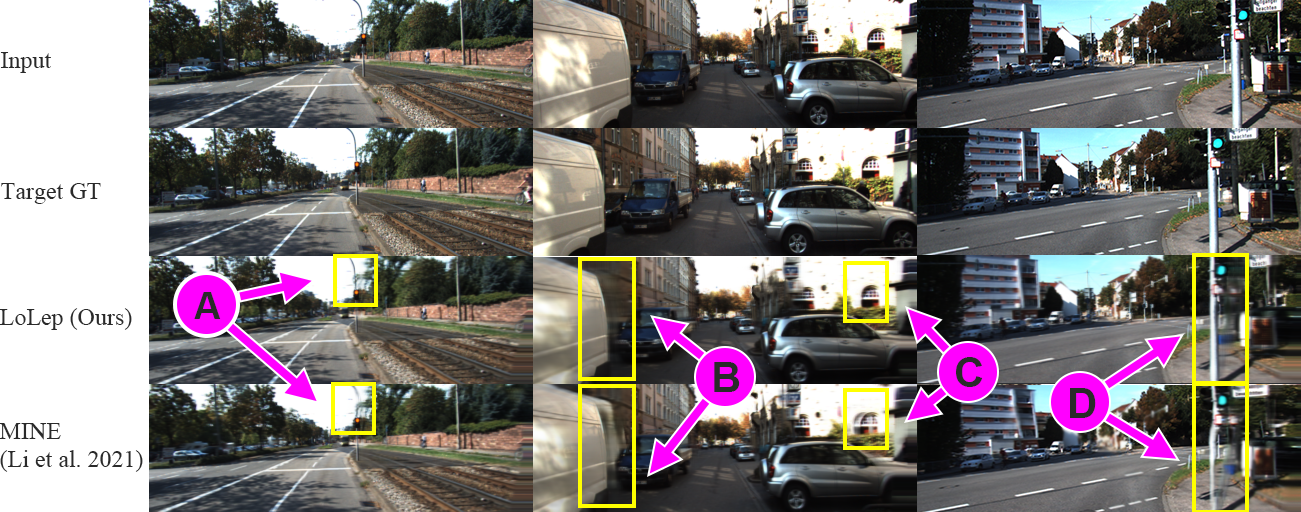}
   \caption{{\bf Qualitative comparison on the KITTI dataset.}
    All images are from the test dataset and highlight the benefits of LoLep.
    (A) MINE synthesizes a broken pole. 
    (B) MINE fails to infer occluded regions, thereby causing ghosting. 
    (C) MINE regresses a suboptimal scene representation, thereby generating ghosting.
    (D) MINE synthesizes a twisted pole due to inconsistent depths of the pole.
    } 
\label{fig: kitti_comparison}
\end{figure*}

\begin{table}[]
    \caption{Evaluation results on the KITTI dataset.
    LoLep obtains the best performance compared to prior methods, and even LoLep with fewer planes uses less memory and generates better results than prior methods with more planes.
    The image resolution is $384\times128$.
    The best is in {\bf bold}, and the second best is \underline{underlined}.
    }
    \centering
    \resizebox{85mm}{!}{
    \begin{tabular}{l|c|c|c|c|c|c}
    \hline
    \multirow{2}{*}{Methods}
    & \multirow{2}{*}{LPIPS\textsuperscript{\ref{foot: lpips}}$\downarrow$} & \multirow{2}{*}{SSIM$\uparrow$} & \multirow{2}{*}{PSNR$\uparrow$} & \multirow{2}{*}{RV$\downarrow$} & Memory(MB)$\downarrow$ & Converge$\downarrow$ \\ 
    & & & & & (train/inference) & Iterations \\ \hline
    LDI~\cite{DBLP:conf/eccv/TulsianiTS18} & - & 0.572 & 16.50 & - & - & - \\
    MPI-32~\cite{DBLP:conf/cvpr/TuckerS20} & - & 0.733 & 19.50 & - & - & - \\
    MINE-16~\cite{DBLP:conf/iccv/LiFSDWL21} & 0.146 & 0.806 & 21.48 & 839.17 & \multicolumn{1}{r|}{\bm{{8495}} / \bm{{2039}}} & \multicolumn{1}{r}{30k} \\
    MINE-32~\cite{DBLP:conf/iccv/LiFSDWL21} & {0.134} & 0.813 & 21.52 & 492.50 & \multicolumn{1}{r|}{14351 / 3167} & \multicolumn{1}{r}{30k} \\
    MINE-64\textsuperscript~\cite{DBLP:conf/iccv/LiFSDWL21} & 0.127 & 0.818 & 21.65 & 197.65 & \multicolumn{1}{r|}{28457 / 5287} & \multicolumn{1}{r}{34k} \\ \hline
    LoLep-16 (full) & 0.134 & 0.820 & 21.88 & 138.74 & \multicolumn{1}{r|}{\underline{10868} / \underline{2232}} & \multicolumn{1}{r}{\bm{{24k}}} \\
    LoLep-32 (full) & \underline{{0.122}} & \underline{{0.825}} & \underline{{22.07}} & \multicolumn{1}{r|}{\underline{{89.61}}} & \multicolumn{1}{r|}{17208  / 3478} & \multicolumn{1}{r}{\underline{25k}} \\
    LoLep-64 (full) & \bm{{0.117}} & \bm{{0.828}} & \bm{{22.17}} & \multicolumn{1}{r|}{\bm{{49.53}}} & \multicolumn{1}{r|}{32421 / 5639} & \multicolumn{1}{r}{{28k}} \\ \hline
    \end{tabular}
    }
    \begin{tablenotes}
      \scriptsize	
      \item * The training batch size is 4 and that of the inference is 1.
    \end{tablenotes}
    \label{tab: kitti_eval}
\end{table}

\subsection{View Synthesis on KITTI}
\label{sec: kitti_eval}

Using the same settings as previous works~\cite{DBLP:conf/cvpr/TuckerS20, DBLP:conf/eccv/TulsianiTS18, DBLP:conf/iccv/LiFSDWL21}, 20 city sequences of the dataset are used to train our models, 4 sequences are used for validation, and the remaining 4 sequences are used for testing.
During training, the left or the right image is randomly taken as the source image, and the other is the target image.
Following~\cite{DBLP:conf/iccv/LiFSDWL21}, we also crop 5\% from all sides of images when testing. 
The quantitative results have been shown in Table \ref{tab: kitti_eval}.
LoLep has better performance than previous methods, and even our models with fewer planes use less memory and generate better results than models of previous methods with more planes (e.g., LoLep-16 vs. MINE-32, MINE-64 and MPI-32, LoLep-32 vs. MINE-64).
The massive reduction of RV shows that our regressed locations allow the volume rendering to concentrate on fewer and more accurate planes, thereby generating sharper results and alleviating artifacts for novel views.
As shown in Figure \ref{fig: kitti_comparison}, LoLep can handle occlusions better (Figure \ref{fig: kitti_comparison}(B)) and generate more reasonable geometry and shaper images (Figure \ref{fig: kitti_comparison}(A), (C)-(D)).

\begin{table}[h]
    \caption{Evaluation results on the RealEstate10K dataset.
    LoLep generates better results than prior methods.
    The image resolution is $384\times256$.
    The best is in {\bf bold}, and the second best is \underline{underlined}.
    }
    \centering
    \resizebox{85mm}{!}{
    \begin{tabular}{l|c|c|c|c|c}
    \hline
    \multirow{2}{*}{Methods}
    & \multirow{2}{*}{LPIPS\textsuperscript{\ref{foot: lpips}}$\downarrow$} & \multirow{2}{*}{SSIM$\uparrow$} & \multirow{2}{*}{PSNR$\uparrow$} & Memory(MB)$\downarrow$ & Converge$\downarrow$ \\ 
    & & & & (train/inference) & Iterations \\ \hline
    SynSin~\cite{DBLP:conf/cvpr/WilesGS020} & - & 0.740 & 22.31 & - & -\\
    MPI-32~\cite{DBLP:conf/cvpr/TuckerS20} & - & 0.785 & 23.52 & - & -\\
    MINE-16~\cite{DBLP:conf/iccv/LiFSDWL21} & 0.208 & 0.804 & 23.71 & \multicolumn{1}{r|}{\bm{{13195}} / \bm{{2671}}} & \multicolumn{1}{r}{1500k}\\
    MINE-32~\cite{DBLP:conf/iccv/LiFSDWL21} & {0.187} & 0.813 & 24.33 & \multicolumn{1}{r|}{19842 / 3955} & \multicolumn{1}{r}{1580k} \\
    MINE-64~\cite{DBLP:conf/iccv/LiFSDWL21} & 0.176 & {0.818} & {24.50} & \multicolumn{1}{r|}{34231 / 6421} & \multicolumn{1}{r}{1660k} \\ \hline
    LoLep-16 (full) & 0.191 & 0.816 & 24.41 & \multicolumn{1}{r|}{\underline{14963} / \underline{2932}} & \multicolumn{1}{r}{\bm{{1000k}}} \\
    % LoLep-32 (full) & & & & \\ \hline
    LoLep-32 (full) & \underline{{0.174}} & \underline{{0.828}} & \underline{{25.02}} & \multicolumn{1}{r|}{22987 / 4413} & \multicolumn{1}{r}{\underline{1030k}} \\
    LoLep-64 (full) & \bm{{0.161}} & \bm{{0.832}} & \bm{{25.14}} & \multicolumn{1}{r|}{38754 / 6845} & \multicolumn{1}{r}{{1100k}} \\ \hline
    \end{tabular}
    }
    \begin{tablenotes}
      \scriptsize	
      \item * The training batch size is 4 and that of the inference is 1.
    \end{tablenotes}
    \label{tab: real_eval}
\end{table}

\subsection{View Synthesis on RealEstate10K}
\label{sec: real_eval}

\begin{figure*}
    \centering
    \includegraphics[scale=0.32]{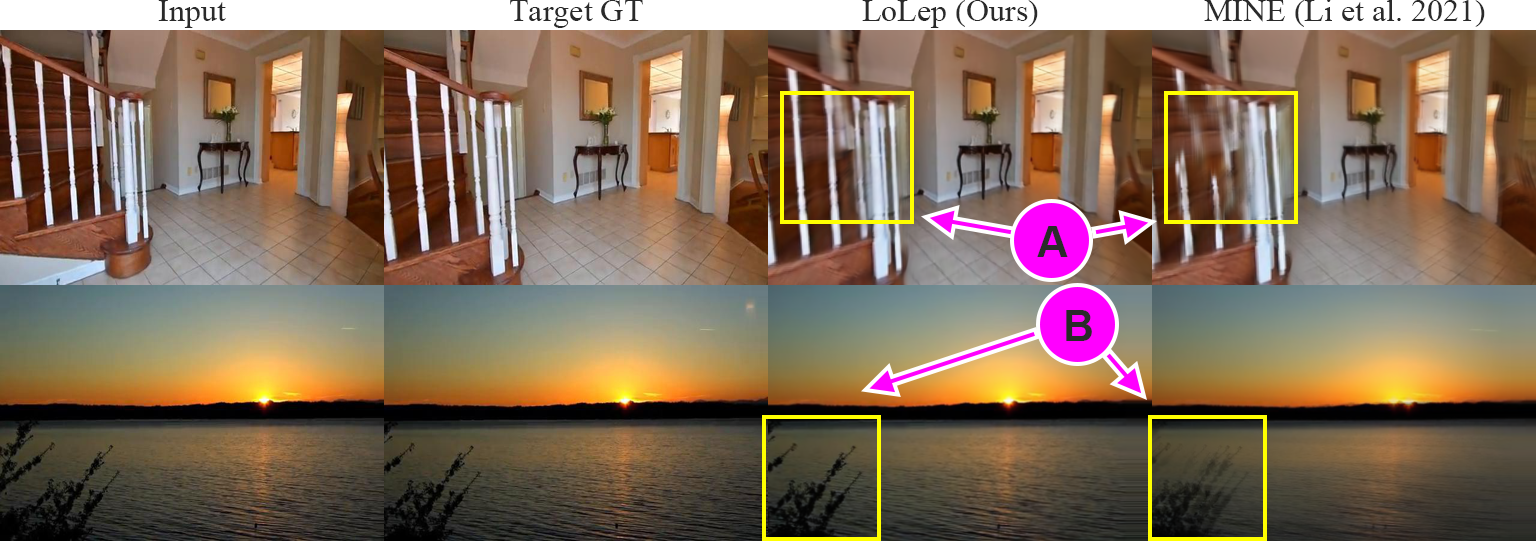}
    \caption{{\bf Qualitative comparison on the RealEstate10K dataset.}
    (A) MINE fails to infer the geometry of the balustrade in stairs.
    (B) MINE generates many artifacts and blurry regions. In contrast, LoLep generates improved results.
    }
    \label{fig: Real_comparison}
\end{figure*}

RealEstate10K~\cite{DBLP:journals/tog/ZhouTFFS18} is a large-scale dataset collected from video clips on YouTube and consists of over 70,000 video sequences.
Since different sequences have different scales, we use COLMAP~\cite{DBLP:conf/cvpr/SchonbergerF16, DBLP:conf/eccv/SchonbergerZFP16} to generate sparse point clouds of each sequence for scale-invariant learning~\cite{DBLP:conf/cvpr/TuckerS20}.
Due to the huge size of the dataset, we randomly select $10\%$ from the official training sequences to train our model. For testing, we 
randomly sample 600 sequences from the official test split and draw 5 frames from each sequence as source images.
During both training and testing, target images are in the same sequence as source images and are randomly selected within 30 frames of source images.
As shown in Table \ref{tab: real_eval}, LoLep generates better results than previous methods on all metrics.
Qualitative results in Figure \ref{fig: Real_comparison} further demonstrate that LoLep synthesizes sharper and more realistic images for novel views (Figure \ref{fig: Real_comparison}(A)-(B)).

\subsection{View Synthesis on Flowers Light Fields}
\label{sec: flowers_eval}

The Flowers Light Fields dataset~\cite{DBLP:conf/iccv/SrinivasanWSRN17}  consists of 3,343 light field photos of flowers.
During training, a random image is selected as the source image and another image in the same light field is taken as the target image. 
In testing, we use a center image as the source image and four corner images as the target images.
The training and testing splits are obtained from~\cite{DBLP:conf/iccv/LiFSDWL21}.
Quantitative results are shown in Table \ref{tab: flowers_eval} and qualitative results are shown in our SMs.

\begin{table}[]
    \caption{Evaluation results on the Flowers Light Field dataset.
    The image resolution is $512\times384$.
    The best is in {\bf bold}, and the second best is \underline{underlined}.
    }
    \centering
    \resizebox{85mm}{!}{
    \begin{tabular}{l|c|c|c|c|c}
    \hline
    \multirow{2}{*}{Methods}
    & \multirow{2}{*}{LPIPS\textsuperscript{\ref{foot: lpips}}$\downarrow$} & \multirow{2}{*}{SSIM$\uparrow$} & \multirow{2}{*}{PSNR$\uparrow$} & Memory(MB)$\downarrow$ & Converge$\downarrow$ \\ 
    & & & & (train/inference) & Iterations \\ \hline
    LLFF~\cite{DBLP:conf/iccv/SrinivasanWSRN17} & - & 0.822 & 28.10 & - & - \\
    MPI-32~\cite{DBLP:conf/cvpr/TuckerS20} & - & 0.851 & 30.10 & - & - \\
    MINE-16~\cite{DBLP:conf/iccv/LiFSDWL21} & 0.208 & 0.862 & 29.86 & \multicolumn{1}{r|}{\bm{{13251}} /\;\enspace\bm{{4042}}} & \multicolumn{1}{r}{250k} \\ 
    MINE-32~\cite{DBLP:conf/iccv/LiFSDWL21} & 0.201 & 0.868 & 30.17 & \multicolumn{1}{r|}{20340 /\;\enspace6373} & \multicolumn{1}{r}{252k} \\
    MINE-64~\cite{DBLP:conf/iccv/LiFSDWL21} & 0.188 & {0.873} & {30.31} & \multicolumn{1}{r|}{34503 / 11407} & \multicolumn{1}{r}{260k} \\ \hline
    LoLep-16 (full) & {0.198} & 0.868 & 30.21 & \multicolumn{1}{r|}{\underline{15746} /\;\enspace\underline{4832}} & \multicolumn{1}{r}{\bm{{200k}}} \\ 
    LoLep-32 (full) & \underline{{0.183}} & \underline{{0.876}} & \underline{{30.35}} & \multicolumn{1}{r|}{23427 /\;\enspace7231} & \multicolumn{1}{r}{\bm{{200k}}} \\ 
    LoLep-64 (full) & \bm{{0.181}} & \bm{{0.880}} & \bm{{30.41}} & \multicolumn{1}{r|}{39054 / 12384} & \multicolumn{1}{r}{\underline{205k}} \\ \hline
    \end{tabular}
    }
    \begin{tablenotes}
      \scriptsize	
      \item * The training batch size is 2 and that of the inference is 1.
    \end{tablenotes}
    \label{tab: flowers_eval}
\end{table}

\begin{table}
    \caption{Depth Evaluation on NYU-Depth V2 and iBims-1.
    The significant improvements show the superiority of LoLep in regressing more accurate scene representation.
    The best is in {\bf bold}.
    }
    \centering
    \resizebox{85mm}{!}{
    \begin{tabular}{c|l|ccccccc}
    \hline
    Data & Methods & rel$\downarrow$ & log10$\downarrow$ & RMS$\downarrow$ & $\sigma 1 \uparrow$ & $\sigma 2\uparrow$ & $\sigma 3 \uparrow$ & RV$\downarrow$ \\ \hline 
    \multirow{2}{*}{NYU} & MINE-64 & 0.17 & 0.07 & 0.58 & 0.77 & 0.93 & 0.98 & 3.41 \\ 
    & LoLep-64 & \bm{{0.15}} & \bm{{0.06}} & \bm{{0.49}} & \bm{{0.81}} & \bm{{0.95}} & \bm{{0.99}} & \bm{{2.53}} \\ \hline
    \multirow{2}{*}{iBims}& MINE-64 & 0.17 & 0.08 & 0.73 & 0.75 & 0.91 & 0.96 & 3.95 \\ 
    & LoLep-64 &\bm{{0.15}} & \bm{{0.06}} & \bm{{0.62}} & \bm{{0.81}} & \bm{{0.94}} & \bm{{0.99}} & \bm{{3.02}} \\ \hline
    \end{tabular}
    }
    \label{tab: depth evaluation}
\end{table}

\subsection{Depth Evaluation on NYU-V2 and iBims-1}

We further perform the depth evaluation on the NYU-Depth V2~\cite{DBLP:conf/eccv/SilbermanHKF12} and iBims-1~\cite{DBLP:conf/eccv/KochLF018} datasets using 64-plane models trained on RealEstate10K; the results are shown in Table \ref{tab: depth evaluation}.
Since the models are trained on RealEstate10K but evaluated on other datasets, the generalization of models is the key to obtaining good performance.
In our settings, we only using 10\% of the dataset for training makes our models not have good generalization ability for new datasets, so the quality of depth maps is not comparable to state-of-the-art methods in depth estimation.
However, depth maps generated by our models are significantly better than those of MINE with the same settings, which demonstrates that our method can regresses more accurate scene representation, a major point of improvement.
Qualitative comparisons are shown in our SMs.

\subsection{Ablation Study}
\label{sec: ablation_study}

To further demonstrate the benefits of our proposed methods, we perform some ablation studies on the KITTI dataset; the results are shown in Table \ref{tab: ablation_study}.
(a.1)-(a.4) compare different approaches to obtaining plane locations. 
(a.1) is our baseline, obtaining plane locations by first dividing the disparity space into N bins and then randomly selecting locations in each bin as ~\cite{DBLP:conf/iccv/LiFSDWL21} does.
(a.2) obtains plane locations by equally dividing the disparity space.
(a.3) learns plane locations globally.
(a.4) learns plane locations locally using our proposed sampler with U-opt, which is the best way verified by our experiments.
(b.1)-(b.2) show the functionality of our proposed components, and (b.3) shows that only using a reprojection loss without an occlusion mask degrades the performance.

We also perform some experiments to explore the benefits of our BS-SA module.
(c) shows that the original self-attention cannot be applied to feature maps of size 32$\times$96 due to the vast memory overhead.
However, our BS-SA module even can be applied to feature maps of size 64$\times$192 for higher performance ((d.1) and (e.3)).
Compared to the original self-attention, our BS-SA module can obtain comparable accuracy with less memory ((d.1)-(d.2)).
In addition, as shown in (e.1)-(e.4), we can trade between the memory overhead and the performance by adjusting the number of sampling points $M$.
As $M$ increases, the improvements of our BS-SA module increase.
However, a too small value of $M$ leads to performance degradation because too few samples cannot guide parameters to update in the right direction.

\section{Discussion on Methods using Monocular Depth Estimators}

\begin{figure}[tb]
    \centering
    \includegraphics[width=\linewidth]{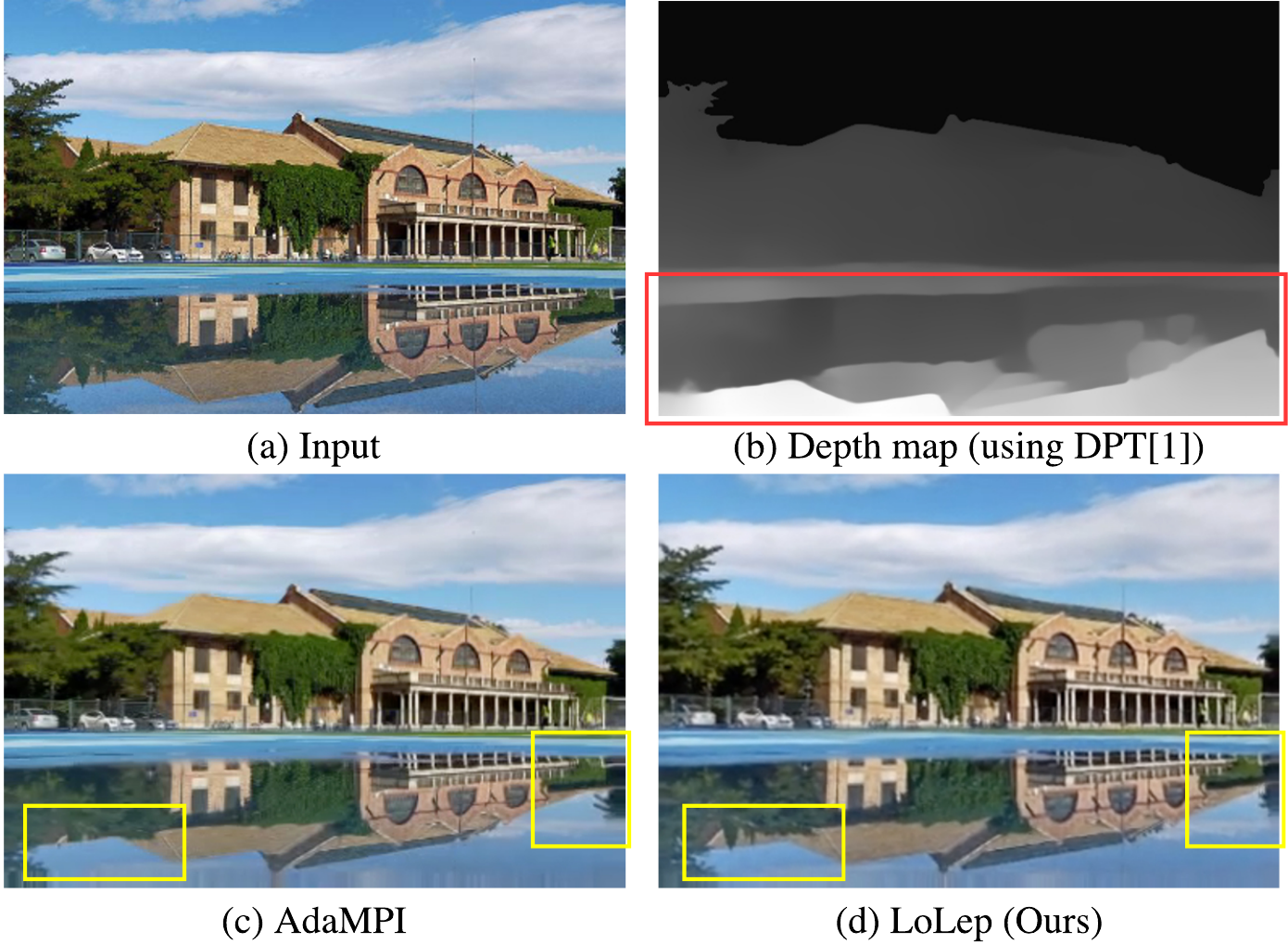}
    \caption{\textbf{A failure case of AdaMPI on a scene with mirror reflections.}
    (b) is generated using DPT~\cite{DBLP:conf/iccv/RanftlBK21}, consistent with AdaMPI.
    }
    \label{fig:reflection_comp}
\end{figure}

In some cases, off-the-shelf monocular depth estimators indeed aid in learning reasonable locations of MPI~\cite{DBLP:conf/cvpr/ShihSKH20, DBLP:journals/tog/LiK20, DBLP:conf/siggraph/HanWY22}.
However, monocular depth estimation is still a challenging problem and has many unsolved limitations~\cite{DBLP:journals/corr/abs-2003-06620} (e.g., reflections and transferability), inevitably introducing these limitations to the single-view view synthesis.
For example, Fig.~\ref{fig:reflection_comp} compares our approach to AdaMPI~\cite{DBLP:conf/siggraph/HanWY22} on a real-world scene with mirror reflections. Due to the wrong depth estimation for reflection regions (the red box), AdaMPI produces obvious artifacts (yellow boxes). 
In contrast, our approach generates more resonable results. 
A possible explanation is that our sampler is jointly learned with the view synthesis task and solve only a simpler optimization problem (learning locations for different depth levels) than monocular depth estimation (learning per-pixel depth values).

\section{Conclusion, Limitations, and Future Work}
\label{sec: conclusion}

We present a novel method, LoLep, for single-view view synthesis that regresses locally-learned planes to represent scenes accurately, thus generating better novel views.
This includes a novel disparity sampler with different parameter optimizing strategies, exploration of an occlusion-aware reprojection loss, and a novel BS-SA module that can be applied to large feature maps.
Results on different datasets and real-world images show that LoLep can generate better results and achieve new state-of-the-art performance.

\noindent
\textbf{Limitations.} 
Although locally-learned planes prevent all planes from clustering around a certain disparity and obtain promising results, it is a suboptimal solution.
An optimal solution should allow planes to be optimized through the whole disparity space and prevent them from clustering using some new techniques.
In the future, we will work on this topic and provide a further solution.

{\small
\bibliographystyle{ieee_fullname}
\bibliography{egbib}
}

\end{document}